\documentclass[11pt,a4paper]{article}
\usepackage[hyperref]{acl2020}
\usepackage{times}
\usepackage{latexsym}

\usepackage{booktabs}
\usepackage{amsmath}
\usepackage{xspace}
\usepackage{amsfonts}
\usepackage{amssymb}
\usepackage{color}

\newcommand{\nop}[1]{}

\newcommand{\expect}[1]{\ensuremath{\underset{#1}{\mathbb{E}}\xspace}}
\DeclareMathOperator*{\argmax}{arg\!max}

\usepackage[disable]{todonotes} % use this to disable notes
\makeatletter
\newcommand*\iftodonotes{\if@todonotes@disabled\expandafter\@secondoftwo\else\expandafter\@firstoftwo\fi}  % defines \iftodonotes{<true>}{<false>}, thanks to https://tex.stackexchange.com/questions/126559/conditional-based-on-packageoption
\makeatother

\newlength{\extramargin}
\usepackage{calc}
\iftodonotes{\usepackage[paperwidth=\paperwidth+\extramargin*2,marginparwidth=\marginparwidth+\extramargin,width=\textwidth,height=\textheight]{geometry}}{}
\newcommand{\dataset}{\textsc{LogicNLG}\xspace}

\usepackage{microtype}
\usepackage{graphicx}
\aclfinalcopy % Uncomment this line for the final submission
\def\aclpaperid{155} %  Enter the acl Paper ID here

\title{Logical Natural Language Generation from Open-Domain Tables}

\author{Wenhu Chen$^{1}$, Jianshu Chen$^{2}$, Yu Su$^{3}$, Zhiyu Chen$^{1}$ and William Yang Wang$^{1}$\\
University of California, Santa Barbara, CA, USA$^{1}$\\
Tencent AI Lab, Bellevue, WA, USA$^{2}$\\
The Ohio State University, Columbus, Ohio, USA$^{3}$\\
\tt{\{wenhuchen, zhiyuchen, william\}@cs.ucsb.edu} \\
\tt{jianshuchen@tencent.com, su.809@osu.edu}
}

\date{}

\begin{document}
\maketitle
\begin{abstract}
Neural natural language generation (NLG) models have recently shown remarkable progress in fluency and coherence. However, existing studies on neural NLG are primarily focused on surface-level realizations with limited emphasis on logical inference, an important aspect of human thinking and language. In this paper, we suggest a new NLG task where a model is tasked with generating natural language statements that can be \emph{logically entailed} by the facts in an open-domain semi-structured table. To facilitate the study of the proposed logical NLG problem, we use the existing TabFact dataset~\cite{chen2019tabfact} featured with a wide range of logical/symbolic inferences as our testbed, and propose new automatic metrics to evaluate the fidelity of generation models w.r.t.\ logical inference. The new task poses challenges to the existing monotonic generation frameworks due to the mismatch between sequence order and logical order. In our experiments, we comprehensively survey different generation architectures (LSTM, Transformer, Pre-Trained LM) trained with different algorithms  (RL, Adversarial Training, Coarse-to-Fine) on the dataset and made following observations: 1) Pre-Trained LM can significantly boost both the fluency and logical fidelity metrics, 2) RL and Adversarial Training are trading fluency for fidelity, 3) Coarse-to-Fine generation can help partially alleviate the fidelity issue while maintaining high language fluency. The code and data are available at \url{https://github.com/wenhuchen/LogicNLG}.
\end{abstract}

\section{Introduction}
\begin{figure}[!t]
    \centering
    \includegraphics[width=1.0\linewidth]{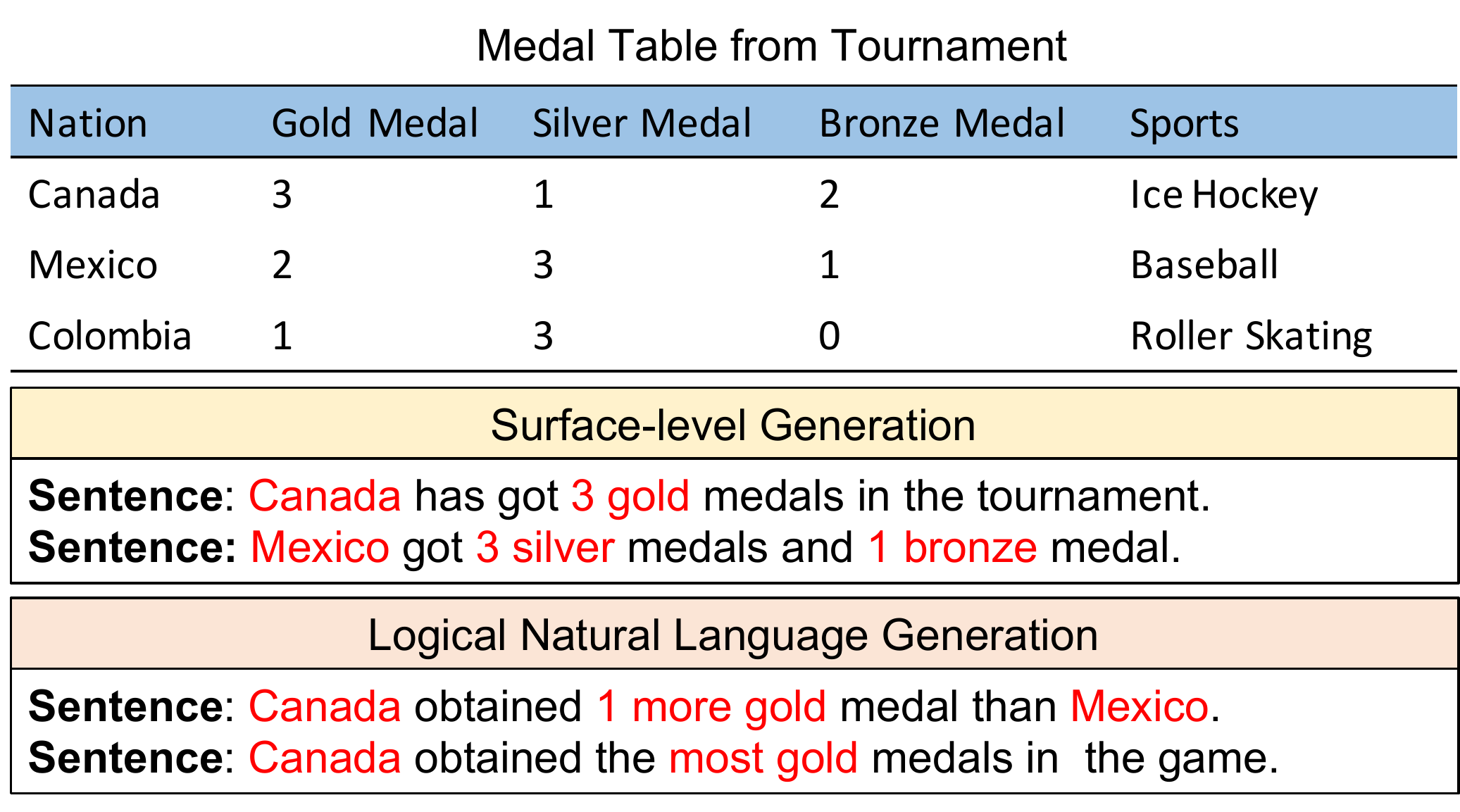}
    \caption{Table-to-text generation examples with and without implicit logical inference. Logical NLG requires a generation model to generate natural language statements that can be logically entailed by the facts in the table instead of simply restating certain superficial facts in natural language. \nop{Keywords  like ``1 more" and ``most" involve logical operations like `Diff' (the difference between numeric values) and `Argmax' (select the row with a maximum value over a column). }}
    \label{fig:examples}
    \vspace{-3ex}
\end{figure}
Neural network models, especially the recent wave of massive models like BERT \cite{devlin2019bert} and GPT-2 \cite{radford2019language}, have shown the ability to generate natural language text at an astonishing level of fluency and coherence. For the generated text to fulfill its purpose, however, a critical property that is necessary but often overlooked is \emph{fidelity}, i.e., what is generated should be faithful to the underlying data, knowledge, or meaning representation.  
A line of recent work has started to address the surface-level fidelity issue of natural language generation (NLG) by encouraging the model to learn to reuse the verbatim of certain inputs through copy mechanism~
\cite{see2017get,gu2016incorporating,wiseman2017challenges,liu2018table}, structured attention~\cite{liu2018table}, or planning and selection/entity modeling~\cite{puduppully2019data,puduppully-etal-2019-data}.
While shown to be effective, most such methods so far are primarily focused on surface-level realization and simply restate the facts in the underlying data (Figure \ref{fig:examples}). 

However, humans have the ability to generalize beyond superficial facts (e.g., ``\emph{Canada has got 3 gold medals.}'') by inferring and communicating with new statements that can be entailed from these facts (e.g., ``\emph{Canada obtained the most gold medals.}''). We believe it is important for NLG models to be able to generalize beyond the superficla facts given to them as well. Therefore, we propose a new task, \emph{logical NLG}, where a model is tasked with generating natural language statements that can be \emph{logically entailed} by the given data (i.e., the \emph{premises}). The new task requires a model to jointly reason and generate sentences that are consistent both linguistically and logically. Since there are a variety of reasoning/inference tasks such as natural language inference~\cite{bowman2015large} and commonsense reasoning~\cite{talmor2019commonsenseqa}, to avoid confusion, this paper is specifically focused on inferences involving symbolic operations over the given table~\cite{pasupat2015compositional}.

To empower research in this direction, we collect a new corpus \dataset based on the existing TabFact~\cite{chen2019tabfact}, which brings two major renovations to the existing NLG paradigm: 1) the text involves diversified types of logical inferences including math operations like max/min/sum/add, comparison operations like same/different, and counting operations like total/only. A more detailed description of logical inference is listed in the Appendix. 2) while existing datasets are often restricted to a specific domain such as weather~\cite{liang2009learning}, restaurant~\cite{dusek2019e2e}, NBA~\cite{wiseman2017challenges}, etc, \dataset uses open-domain tables without prior knowledge about their schema. As such, existing methods based on surface-level copying~\cite{see2017get,gu2016incorporating,puduppully2019data} becomes insufficient, so are the existing fidelity evaluation based on the surface-level information extraction~\cite{wiseman2017challenges,rohrbach2018object,dhingra2019handling}, which extracts surface triples in a certain pre-defined form (i.e. subj-pred-obj, n-gram) and compare them with the surface content given in the knowledge. %As demonstrated in~\autoref{fig:definition}, the IE-based evaluation metric often fails to grasp the semantics of the sentence or yields false negatives.

\begin{figure}[!t]
    \centering
    \includegraphics[width=1.0\linewidth]{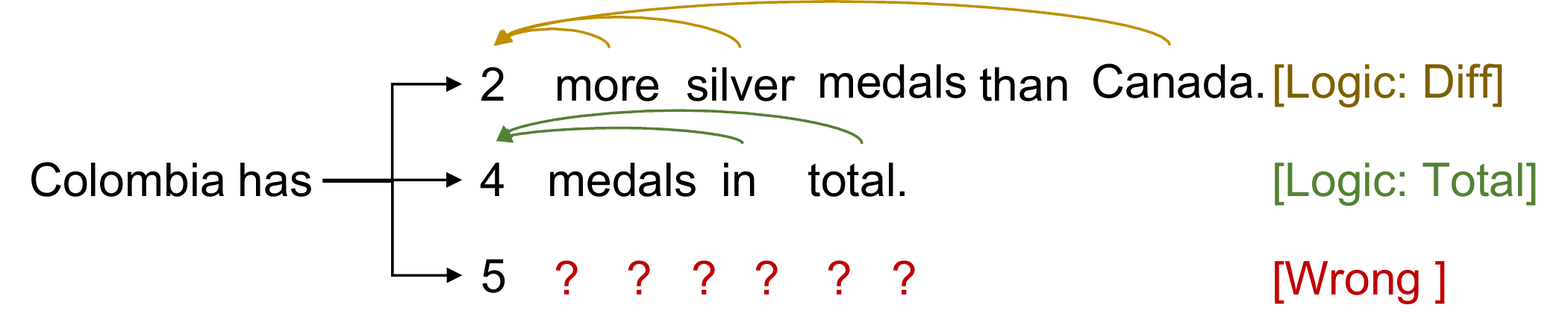}
    \caption{When making the decision at the third step, the model needs to foresee the future tokens to ensure logical consistency. There is no back-tracking once the model makes a wrong decision like ``5''.}
    \label{fig:backward}
    \vspace{-3ex}
\end{figure}
Most neural generation models follow a monotonic generation schema from left to right with the current prediction only depending on the preceding words. Logical NLG poses unique challenges to the traditional generation scheme due to the mismatch between \textit{sequence order} and \textit{logical order}. As illustrated in~\autoref{fig:backward}, the word ``2'' is derived from the logical inference of `diff(Silver medal of Colombia, Silver medal of Canada)) $\rightarrow$ 2.' In other words, the logical order of word ``2'' should be after ``more'', ``silver'', and ``Canada'', while the sequence order of ``2'' is before those words. Since the monotonic generation scheme is purely based on sequence order while agnostic to logical order, existing NLG models struggle to maintain the fidelity as they cannot model the logical dependency on future tokens. To alleviate such an order mismatch, an NLG model must have the capability to plan ahead for the next few steps before generation. In this context, we believe \dataset to be an important testbed to study such a planing/inference ability in generation models~\cite{ford2018importance,welleck2019non}. In this paper, we further propose a non-monotonic coarse-to-fine generation model and show that it is able to alleviate the order mismatch problem and achieve better performance. The contribution of this work is three-fold: 
\vspace{1ex}\\
\indent $i$) We propose a new research problem of logical natural language generation, and provide novel metrics to approximately evaluate the logical fidelity of generation models.\\
\indent $ii$) We justify the mismatch problem between sequence order and logical order of the traditional monotonic generation scheme in logical NLG.\\
\indent $iii$) We conduct comprehensive experiments with state-of-the-art neural generation models under both automatic and human evaluation, which demonstrates the challenges and opportunities for future research on logic NLG.\\

\begin{table*}[!t]
\small
\centering
\begin{tabular}{lllllllll} 
\toprule
           & Vocab & Examples & Vocab/Sent  & Tables & Domain    & Source    & Inference     & Schema             \\ 
\midrule
WEATHERGOV & 394   & 22.1K  & 0.01 & 22.1K  & Weather   & Crawled   & No        & Known      \\
WikiBIO    & 400K  & 728K   & 0.54 & 728K   & Biography & Crawled   & No        & Limited  \\
ROTOWIRE   & 11.3K & 4.9K   & 0.72  & 4.9K   & NBA      & Annotated & Few       & Known      \\
\dataset & 122K  & 37.0K  & \textbf{3.31}  & 7.3K   & \textbf{Open}      & Annotated & \textbf{Rich}      & \textbf{Unlimited}  \\
\bottomrule
\end{tabular}
\caption{Comparison of \dataset against existing NLG datasets in different aspects.}
\label{tab:stats}
\vspace{-1ex}
\end{table*}

\section{Dataset and Problem Definition}
\label{sec:pro}
Existing NLG datasets~\cite{chen2008learning,dusek2019e2e,lebret2016neural,liang2009learning} are mainly composed of surface-level description over the given records. Though ROTOWIRE~\cite{wiseman2017challenges} involves sporadic inference in the long document, and the inference is restricted to domain-specific knowledge (e.g. double-double, smash, triple-double and other NBA-related terms). Hence, we need a better testbed for studying the proposed problem. 

\begin{figure*}[!t]
    \centering
    \includegraphics[width=0.95\linewidth]{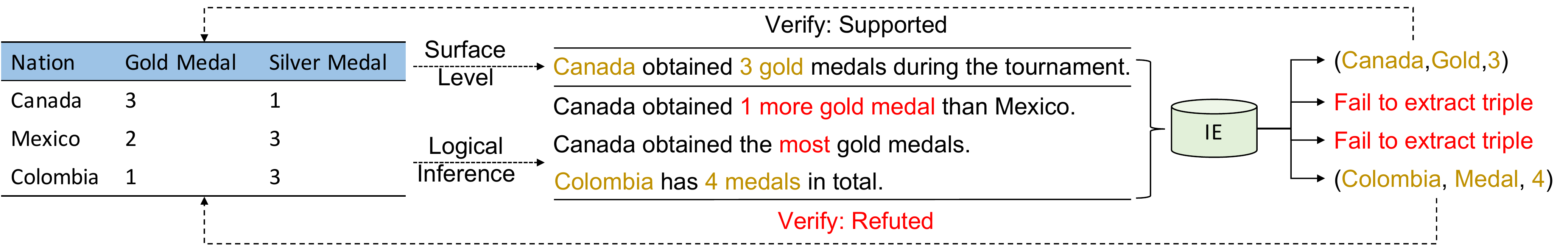}
    \caption{Evaluation of surface-level generation vs. logical natural language generation. It suffices to use IE-based evaluation~\cite{wiseman2017challenges,rohrbach2018object} to verify surface-level generation, but it causes either ``empty triple" or ``false negative" problems to verify logical NLG.}
    \label{fig:definition}
    \vspace{-2ex}
\end{figure*}
\paragraph{Statistics}
We construct a dataset based on TabFact~\cite{chen2019tabfact}, which is a table-based fact-checking dataset with rich logical inferences in the annotated statements. Specifically, we took their positive statements (the sentences which are entailed by the knowledge in the table) collected from ``complex channel" (required to annotate sentences with logical inference) as our target text. To prevent confusion with the original dataset, we name this table-to-text dataset as \dataset, which contains 28,450 training, 4,260 validation and 4,305 test examples based on 7,392 open-domain tables crawled from Wikipedia. Each table has 5 different examples covering diverse types of logical inference. More detailed statistics and comparisons are listed in~\autoref{tab:stats}. \dataset is distinguished from the existing datasets due to:\vspace{1ex}\\
\indent $i$) It involves very rich logical inference, every annotated sentence involves certain types of inference with minimum domain-specific knowledge. The open-domain characteristic simulates a realistic setting, where we cannot enumerate the possible inference based on the scheme, which poses great challenges to the model's generalization capability.\\
\indent $ii$) It is mainly composed of short sentences with an average length of 11 and a simple syntactic structure, which isolates from other linguistic complexity to focus on the problem of logical inference.

\begin{figure}[!t]
    \centering
    \includegraphics[width=0.95\linewidth]{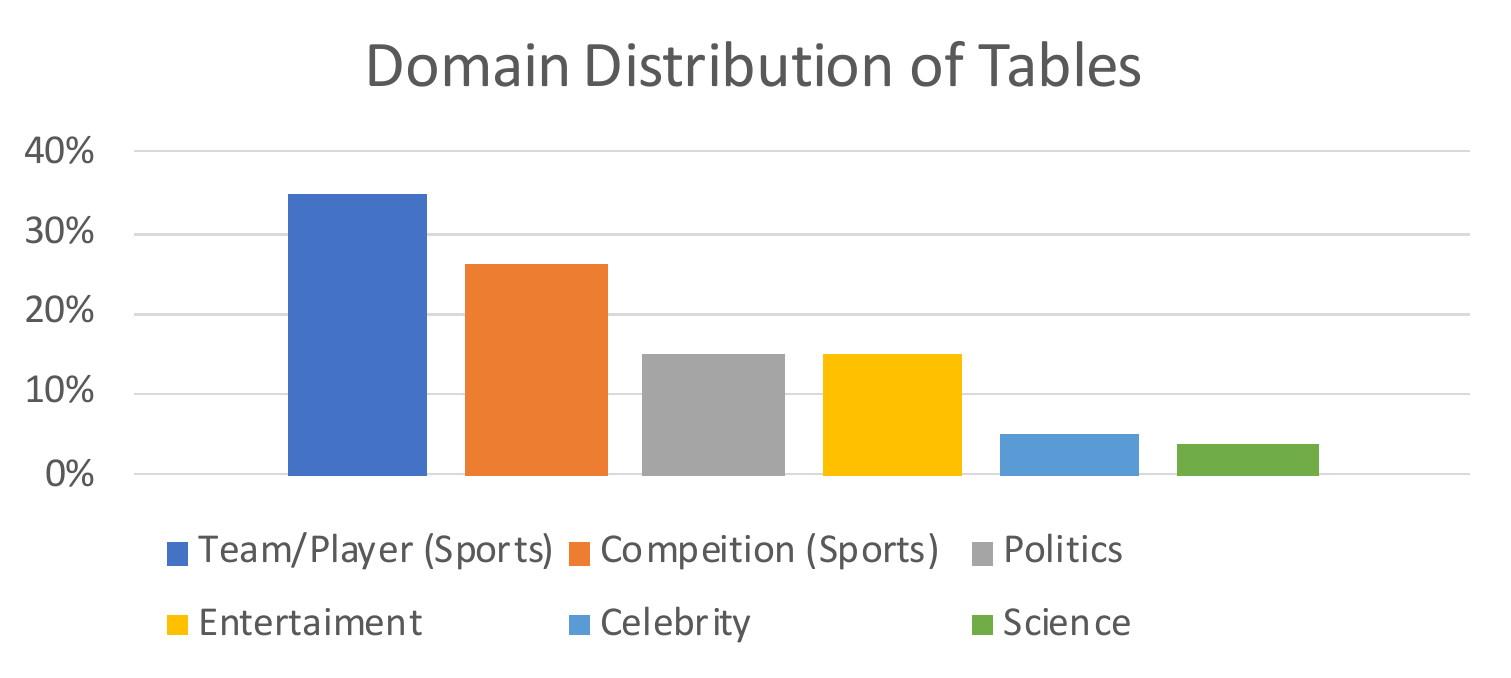}
    \caption{The domain distribution of \dataset. }
    \vspace{-1ex}
    \label{fig:domain}
    \vspace{-2ex}
\end{figure}

The dataset contains tables with open schema crawled from diversified domains~\autoref{fig:domain}. The major categories are sports, politics, and entertainment. %The sports category is mainly divided into two types: 1) the record summary of a player/team,  2) leaderboard of a given competition/season. The politics tables are mainly covering the election-related statistics and the entertainment tables are mainly discussing music, films, etc. 
The schema diversity of the tables make the rule-based system infeasible to apply. Besides, most of the tables have very rich numeral records, which provide a great testbed for logical inference.  

\paragraph{Problem Definition}
Here, we formally define our proposed table-to-text generation task. The input is a table $\textbf{T}$ with its title denoted as a natural language sequence $W$. The table \textbf{T} = $\{T_{i,j} |i \leq R_T , j \leq C_T \}$ has $R_T$ rows and $C_T$ columns with the $T_{ij}$ being the content in the $(i, j)$-th cell. $T_{ij}$ could be a word, a number, a phrase or even a natural language sentence. The annotated statement is a sentence $Y=y_1, y_2, \cdots, y_n$, we aim to train a neural generation model $p(Y|\textbf{T})$ to generate statement $\hat{Y}$ which are both fluent and logically (numerically) supported by the given table $\textbf{T}$.

\section{Automatic Evaluation}
\label{sec:metric}
\begin{figure*}[!t]
    \centering
    \includegraphics[width=0.9\linewidth]{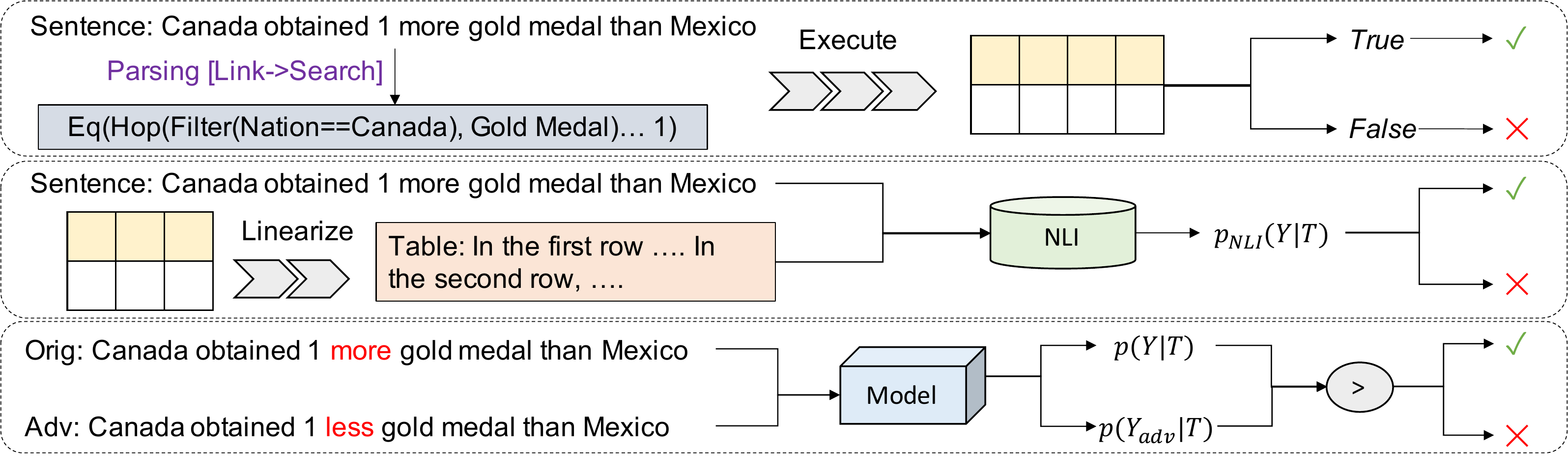}
    \caption{The parsing-based and adversarial evaluation to measure model's correctness in logical reasoning.}
    \label{fig:evaluation}
    \vspace{-2ex}
\end{figure*}
In this section, we discuss the evaluation of our proposed NLG task. The fluency evaluation is simply based on the standard metrics like Perplexity~\cite{bengio2003neural} and BLEU-1,2,3~\cite{papineni2002bleu} based on NLTK~\cite{bird2006nltk}. The most challenging problem is to evaluate the \emph{logical fidelity} of the generated sentences, which is also the core problem of our paper. The existing IE-based extractive evaluation~\cite{wiseman2017challenges}  leads to two issues as shown in~\autoref{fig:definition}: 1) Empty Extraction: the sentence can not be formulated as (subject, predicate, object) structure, thus the IE system fail to extract triples for verification. 2) False Negative: the sentence is a logical composition (instead of surface form) of the fact from the table, the IE system cannot match it against the table. For these reasons, we test two approximate automatic metrics:

\paragraph{Parsing-based Evaluation}
We first propose a model-based evaluation method, which aims to directly extract the meaning representation from the generated sentence and execute it against the table to verify its correctness. Our evaluation is based on weakly-supervised semantic parsing~\cite{liang2009learning,liang2013learning}, the basic idea is to first link entities and predicates in the sentence, and then use linked entities to perform a breadth-first search to synthesize potential logical forms, finally, a scorer is used to re-rank these logical forms and filter out spurious ones. The logical form returns a binary value of \texttt{True} to indicate whether its logic is supported by the knowledge. The basic idea is shown in the upper part of~\autoref{fig:evaluation}, the implementation details are in the Appendix. We pre-train the semantic parser $f_{\gamma}$ on the training set $(\textbf{T}, Y) \in D_{train}$ with weakly supervised algorithm, at test time, we use it to parse a sentence $Y$ into a set of logical forms, which is re-ranked to obtain the highest logical form $P_{best}$. We compute the ratio of $P_{best}$ returning ``true" on $D_{test}$ to approximate model's fidelity.
\begin{align*}
\small
\begin{split}
    \text{SP-Acc} = \expect{(\mathbf{T}, \hat{Y}) \in D_{test}} \mathbb{I}(P_{best} \rightarrow True | P_{best} = f_{\gamma}(\hat{Y}))
    %P_{best} = \argmax_{P \in \mathbf{\hat{P}}} f_{\gamma}(P, \hat{Y}) 
\end{split}
\end{align*}
where $\mathbb{I}$ is the indicator function.
%\begin{figure*}[!t]
%    \centering
%    \includegraphics[width=0.96\linewidth]{evaluation_SP.pdf}
%    \caption{The automatic evaluation by semantic parsing model.}
%    \label{fig:evaluation_sp}
%\end{figure*}
\paragraph{NLI-based Evaluation}
We then propose another model-based evaluation method to complement the parsing-based evaluation (which is sensitive to semantic variation), the basic idea follows~\cite{kryscinski2019evaluating} to evaluate the entailment score between the table and the generated sentence. The NLI model is based on TableBERT~\cite{chen2019tabfact}, which linearizes the table into textual form and uses it as the evidence for natural language inference. The model is trained with TabFact~\cite{chen2019tabfact} dataset containing both positive/negative samples. During the evaluation, we use this NLI model to predict the entailment relationship based on the likelihood of $p_{NLI}(Y|T)$. Finally, we compute the ratio of ``entailed" to approximate model's fidelity:
\begin{align*}
\small
\begin{split}
    \text{NLI-Acc} = \expect{(\mathbf{T}, \hat{Y}) \in D_{test}} \mathbb{I}(p_{NLI}(Y|\textbf{T}) > 0.5)
    %P_{best} = \argmax_{P \in \mathbf{\hat{P}}} f_{\gamma}(P, \hat{Y}) 
\end{split}
\end{align*}
where $\mathbb{I}$ is the indicator function.
\paragraph{Adversarial Evaluation}
Adversarial evaluation~\cite{goodfellow2014explaining,kannan2017adversarial} is used to study the generation model's robustness in logical reasoning. Specifically, we hire human workers from Amazon Mechanical Turk\footnote{\url{https://www.mturk.com/}} to annotate adversarial examples for the test/validation set by simply changing minimum words to revert the logic of the sentence. Such adversarial examples preserve linguistic components like length and style except the logic-related words to specifically disentangle the generation model's reasoning skill. As drawn in the lower part of~\autoref{fig:evaluation}, the original sentence modifies its word ``more" into ``less" as an adversarial example. There are two principles the workers need to follow to make their jobs accepted: 1) the modified words/phrases should be roughly equally frequent to balance the language prior, for example, the number ``1" is better swapped with ``2,3" rather than ``9999" which rarely appears in the corpus. 2) the perturbation should be diverse enough to cover different aspects of logical reasoning skills. We use the generation model $p(Y|\textbf{T};\beta)$ to score the original sentence $Y$ and the adversarial sentence $Y_{adv}$. If the confidence of the original example is higher than its adversarial counterpart, we count it as a successful defense, otherwise as a failed defense. We use the success rate to approximate model's logical reasoning capability.
\begin{align*}
\small
\begin{split}
    \text{Adv-Acc} = \expect{(\textbf{T}, Y, Y_{adv}) \in D_{test}} [\mathbb{I}(p(Y|\textbf{T}) > p(Y_{adv}|\textbf{T}))]    
\end{split}
\end{align*}
where $\mathbb{I}$ is the indicator function. 

\paragraph{Discussion}
Both types of metrics have pros and cons, the SP-Acc and NLI-Acc are two metrics unbiased as it measures the peak samples in the model's likelihood, however, both metrics are based on imperfect models and thus their evaluation scores are inaccurate. SP-Acc is more sensitive to number/calculation errors, and NLI-Acc is more sensitive to semantic errors, therefore, we report both of them to help increase the metrics' robustness. In contrast, the adversarial evaluation score is accurate in terms of reflecting the model's reasoning capability on the given samples. However, as the provided samples might not lie in the high-confidence area of the model's distribution, it is biased in reflecting the model's general reasoning capability. Though these fidelity metric models are prone to errors, in~\autoref{sec:exp}, we show their consistency with human judgment, which reveals their potential to assist human evaluation.

\section{Baselines}
\label{sec:pre}
In this section, we design comprehensive baseline models to perform logical NLG. Specifically, we consider the following two cases: non-pretrained models (LSTM/Transformer) with copy mechanism and pre-trained models (GPT-2 and BERT) with sub-word unit. We train these models with three different algorithms: Maximum Likelihood, Adversarial Training, and Reinforcement Learning.

\begin{figure*}[!t]
    \centering
    \includegraphics[width=1.0\linewidth]{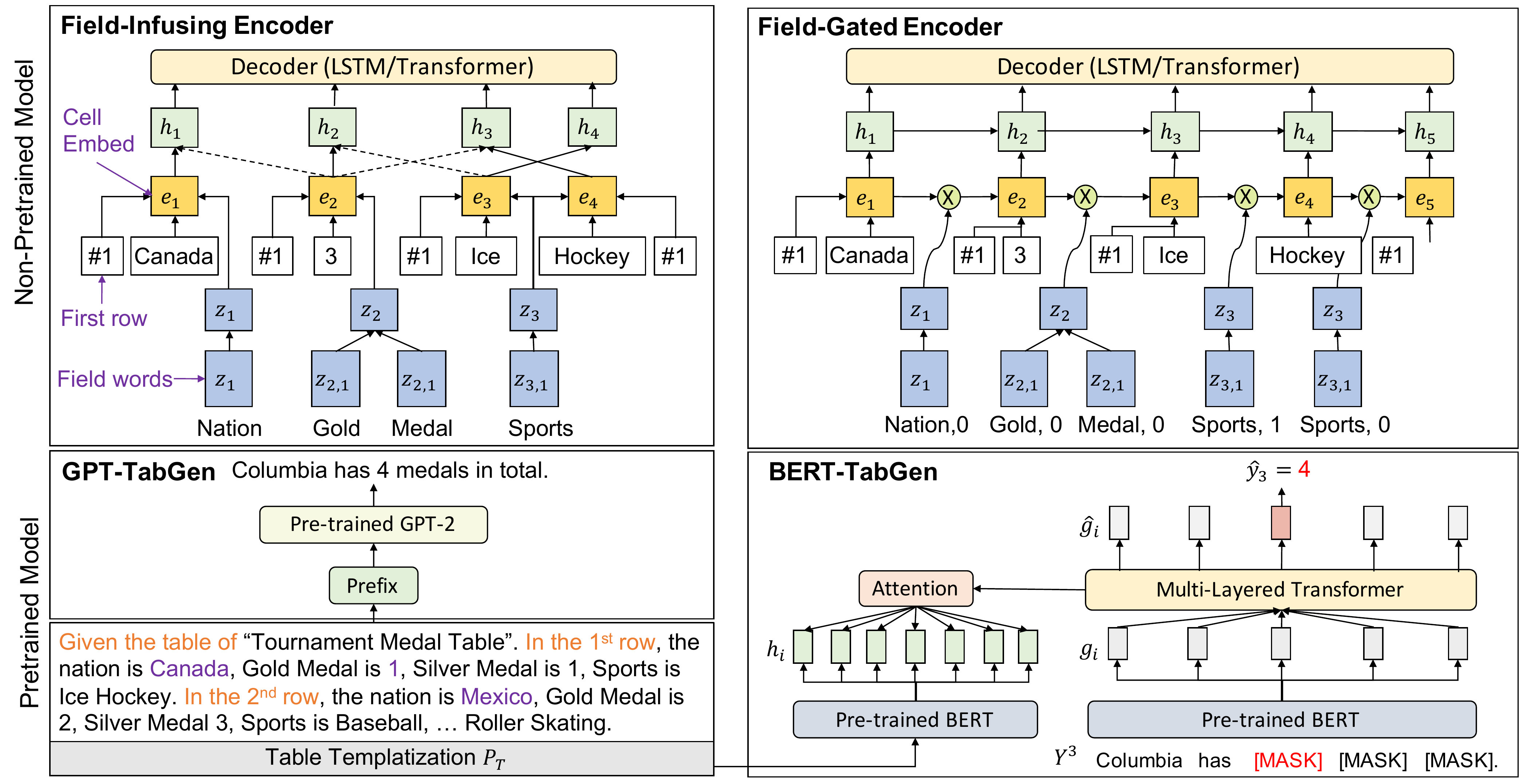}
    \vspace{-4ex}
    \caption{The Non-pretrained and Pre-trained generation models, the detailed table is shown in~\autoref{fig:examples}. }
    \label{fig:baseline}
    \vspace{-2ex}
\end{figure*}

\subsection{Non-pretrained Models}
Here we mainly consider two table encoding methods, namely field-infusing and field-gating. These two methods differ in their strategies to coalesce the field information into cells. After the table is represented as a sequence of vectors, a decoder based on LSTM~\cite{hochreiter1997long} or Transformer~\cite{vaswani2017attention} is applied to generate text token by token. The two methods are depicted in the upper part of~\autoref{fig:baseline}:
\paragraph{Field-Infusing} This strategy is inspired by~\citet{lebret2016neural}. We first use an LSTM~\cite{hochreiter1997long} to encode the table field text word by word and then use the last output $\mathbf{z_i}$ as field representation. This representation is concatenated with the embedding of row index $\#j$ and word embedding at each cell to obtain a position-aware cell embedding $\mathbf{e_k}$ for each word inside the cell. We stack transformers layers on top of the cell embedding to obtain the table representation as $\mathbf{h_i} \in \mathbb{R}^D$ with $D$ as the dimension.
\paragraph{Field-Gating} This strategy is inspired by by~\citet{liu2018table}. Like the previous strategy, we first use an LSTM~\cite{hochreiter1997long} to obtain field representation $\mathbf{z_i}$. The field representation is concatenated with ending distance information as the input to an additional field gate built inside the LSTM as suggested in~\citet{liu2018table}, such a field gate is used to control whether the current cell is already encoded. %For example, the cell ``ice hockey" has two words, in the first step ``sports, 1" will be fed into the encoder to tell LSTM that there will be one more incoming word for this field. While in the second step, ``sports, 0" will be fed in to tell LSTM that the current cell is already finished. 
Such a mechanism can help LSTM to identify the boundary between different cells to grasp local information. 

\subsection{Pre-trained Models}
To further enhance the fluency and resolve the out-of-vocabulary problem, we use pre-trained language models and finetune them on \dataset. Specifically, we consider two models based on GPT-2~\cite{radford2019language} and BERT~\cite{devlin2019bert}, respectively, and name them as GPT-TableGen and BERT-TableGen.
\paragraph{Table Linearization}
We follow previous work on linearizing knowledge base as natural language~\cite{liu2019knowledge,zhang2019relation} to propose ``table linearization", which uses template to flatten the table $\textbf{T}$ as a document $P_T = w_1, \cdots, w_{|T|}$ fed into pre-trained language models to generate statement $Y$, where we use $w_{i}$ to denote the $i$-th word in the generated paragraph $P_T$ and $|T|$ to denote the length of the paragraph (the word $w_i$ is either a table entry or a functional word in the template). As depicted in the left bottom part of~\autoref{fig:baseline}, the original table \textbf{T} is transformed into a paragraph by horizontally scanning each cell $T_{11} \rightarrow T_{1,C_T} \rightarrow T_{R_T, C_T}$ in the table.
\paragraph{GPT-TabGen} we directly feed the paragraph $P_T$ as the input to the pre-trained GPT-2 model and generate the output sentence $Y$. We finetune the model on~\dataset by maximizing the likelihood of $p(Y|P_T;\beta)$, with $\beta$ denoting the parameters of GPT-2 model~\cite{radford2019language}. 
\paragraph{BERT-TabGen} 1) we encode the linearized paragraph $P_T$ using the pre-trained BERT model into the source representation $\mathbf{h_1}, \cdots, \mathbf{h_{|T|}}$. 2) at the $i$-th time step, we replace all the words in the groundtruth statement $Y$ after $i$-th time step by $<$MASK$>$ token and use BERT to encode the partially masked $Y^i$ as $\mathbf{g^i_1}, \cdots, \mathbf{g^i_n}$. 3) we use an attention layer $f_{\theta}$ to obtain the output hidden states $\mathbf{\hat{g}^i_1}, \cdots, \mathbf{\hat{g}^i_{n}}$, where $\mathbf{\hat{g}^i_i}$ is used to predict the word $\hat{y}_i$. We jointly optimize $\beta$ of BERT and $\theta$ to maximize the likelihood of generating text $Y$ conditioned on the table and the masked partial sentence. %as:
%\begin{equation}
%    p(Y|\textbf{T};\theta;\beta) = \prod_{i} p(y_i |f_{\theta}(\mathbf{h_{1:|T|}}, \mathbf{g^i_{1:n}}))
%\end{equation}
As BERT is a bidirectional model, we need to re-encode the target sentence at each step to get $\mathbf{g^i_{1:n}}$. Therefore, the generation is finished with $n$ passes.

\begin{figure*}[!t]
    \centering
    \includegraphics[width=0.90\linewidth]{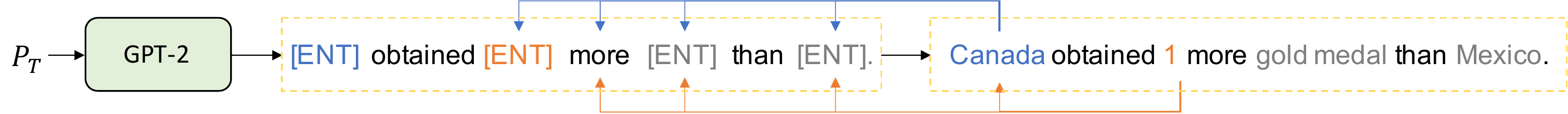}
    \caption{Coarse-to-fine generation scheme: first generates a template, and then realize the surface form. It exposes more context to the surface realization model for better capturing logical dependency. }
    \vspace{-1ex}
    \label{fig:model}
\end{figure*}

\subsection{Training}
Except for the standard maximum likelihood training, we also use the following training algorithms:
\paragraph{Adversarial Regularization}
To encourage the model to ground on the table rather than relying on artificial language priors~\cite{ramakrishnan2018overcoming}, we use an adversarial regularization to enhance the maximum likelihood training. Specifically, we first perform entity resolution to locate all the numbers, count, entities in the sentence and then randomly replace them with entities or numbers appearing in the table $\textbf{T}$. These perturbed samples $Y_{adv}$ are used as adversarial examples to regularize the model's behavior. Formally, we optimize $\beta$ to maximize the objective: 
\begin{align*}
\small
\begin{split}
   \argmax_{\beta} \log p(Y|\mathbf{T};\beta) - \lambda \log p(Y_{adv}|\mathbf{T};\beta)
\end{split}
\end{align*}
where $\lambda$ is the controlling hyper-parameter.
\paragraph{Reinforcement Learning}
The maximum likelihood training is a fluency-driven objective, which is inconsistent with the goal of logical consistency. To bridge the gap, we view the generation problem from the reinforcement learning perspective to optimize the long-term fidelity. We use the trained semantic parser to assign reward to the policy $p(y_i|y_{1:i-1};\beta)$. At $i$-th step, the generator will sample different actions $y_i$ and roll-out from $i+1$-th step to produce a full sequence starting from $y_i$ using greedy search. The full sentence receives a binary score $r(Y, \mathbf{T})$ from the semantic parser as reward. Formally, we optimize the objective:
\begin{align*}
\small
\begin{split}
    \argmax_{\beta} \expect{y_i \sim p(y_i|y_{1:i-1})} [\expect{y_{i+1: n}} [r(y_{1:n}, \mathbf{T})]] \log p(y_i|y_{1:i-1};\beta)
\end{split}
\end{align*}
where we only use one trajectory to approximate the inner roll-out expectation for efficiency. 

\section{Coarse-to-Fine Generation}
\label{sec:model}

As discussed before, the baseline models follow the monotonic generation scheme and suffer from the mismatch between sequence order and logical order (Figure~\ref{fig:backward}). In this section, we propose an imperfect remedy for such a situation based on the coarse-to-fine generation paradigm. 

Before plunging into technical details, it is helpful to first realize the resemblance between logical NLG and semantic parsing~\cite{dong2018coarse}. Compared to traditional NLG tasks like machine translation and summarization, logical NLG is closer to semantic parsing in the sense that a model may make catastrophic errors that are impossible to be corrected at later steps (Figure~\ref{fig:backward}). Therefore, we take inspiration from semantic parsing models~\cite{dong2018coarse} that have proven effective in mitigating such errors and propose a coarse-to-fine generation scheme. We break down generation into two phases. In the first phase, the model only generates a template which determines the global logical structure, while in the second phase the model generates the final, grounded sentence conditioned on the template generated in the first phase. As depicted in~\autoref{fig:model}, we use the entity linker (Section~\ref{sec:metric}) to identify the entities and numbers in the original sentence $Y$ and replace them with placeholder ``[ENT]", which we call as the template $Y_T$. During the generation of GPT-TabGen, instead of directly predicting the final sentence $Y$, we first predict the template $Y_T$ and then $Y$. The process is simply realized by maximizing the overall likelihood of $p(\tilde{Y}|\textbf{T};\beta)$, where $\tilde{Y} = [Y_T; \text{[SEP]}; Y]$. 

Unlike template-based or delexicalized generation~\cite{reiter1997building,wen2015semantically}, which uses rigid slot filling prone to grammatic errors, our fine-grained generation has the flexibility to modify the surface form of non-slot words, which alleviates the linguistic coherence problem~\cite{sharma2016natural}. 

By decoupling sentence structure generation and entity grounding, our proposed coarse-to-fine scheme could partially alleviate the mismatch problem. For example, the generation of ``Canada" is now aware of ``more than" in the latter part of the sentence, which exposes the model to more context than standard monotonic models to help make logically consistent decisions though the dependency on the ``1" and ``Mexico" is still not captured. The proposed two-step generation could be viewed as the first step towards a fully non-monotonic generation model to solve such mismatch problem.

%The grounding part is driven by a neural model rather than logic rules, which cannot guarantee the logical correctness.

\begin{table*}[!t]
\small
\centering
\begin{tabular}{lcccccccccc}
\toprule
Model     & Training & PPL & BLEU-1 & BLEU-2 & BLEU-3 & SP-Acc & NLI-Acc & Adv-Acc \\
\midrule
Field-Gating   + LSTM        & MLE              & 27.7  & 42.3  &  19.5  & 6.9   &  38.0 & 56.8  & 56.2  \\
Field-Gating   + Trans & MLE              & 26.8  & 44.1  &  20.9  & 8.3   &  38.5 &  57.3 & 58.1  \\
Field-Infusing + LSTM        & MLE              & 27.9  & 43.1  &  19.7  & 7.1   & 38.6  & 57.1 & 56.9  \\
Field-Infusing + Trans & MLE              & 26.9  & 43.7  &  20.9   & 8.4  & 38.9  & 57.3 &  58.2   \\ 
\midrule
BERT-TabGen (sm)   & MLE          & 7.5  & 47.8  & 26.3               & 11.9      & 42.2  & 68.1  & 62.4 \\  
GPT-TabGen (sm)  & MLE            & 8.8  & \textbf{48.8}  & \textbf{27.1}   & 12.6        & 42.1 & 68.7  & 62.3  \\
GPT-TabGen (sm) & Adv-Reg         & 12.1  & 45.8  & 23.1              & 9.6             & 40.9  &  68.5  &  64.7 \\
GPT-TabGen (sm) & RL              & 11.3  & 45.1  & 23.6              & 9.1               & \textbf{43.1}  &  67.7 & 61.9  \\
GPT-Coarse-to-Fine (sm) & MLE & - & 46.6 & 26.8 & \textbf{13.3} & 42.7  & \textbf{72.2}  & \textbf{64.9}  \\
\midrule
BERT-TabGen (lg)   & MLE           & 6.3  & 49.1  & 27.7      & 13.5       &  44.4  & 73.9  & 64.0  \\
GPT-TabGen (med) & MLE             & 6.8  & \textbf{49.6}  &  \textbf{28.2} & 14.2 & 44.7  & 74.6  &  64.3 \\
GPT-TabGen (med) & Adv-Reg         & 10.1  & 47.2  & 24.0              & 10.8            & 44.1  &  73.0 & 65.4 \\
GPT-TabGen (med) & RL              & 10.0  & 46.4  & 24.1              & 10.0            & \textbf{45.5}  & 73.3  & 63.7 \\
GPT-Coarse-to-Fine (med)           & MLE & - & 49.0 & 28.3 & \textbf{14.6} & 45.3  & \textbf{76.4} & \textbf{66.0} \\
\bottomrule
\end{tabular}
\caption{The experimental results of different models on the test split of \dataset, where we split the table into non-pretrained LSTM/Transformer, small pre-trained LM (sm) and medium/large pre-trained LM (med/lg). }
\label{tab:un_pretrained}
\vspace{-2ex}
\end{table*}

\section{Experiments}
\label{sec:exp}
In this section, we explain the experimental details and then comprehensively report the automatic evaluation of different generation models and training algorithms. Finally, we will conduct detailed human evaluation and error analysis. 
\subsection{Experiment Setup}
For the non-pretrained models, we fix the hidden size of both LSTM and transformer to be 256, the transformer is 3-layered with 4 heads, while LSTM is also 3-layered. We use Adam optimizer~\cite{kingma2014adam} with a learning rate of 2e-4 to jointly optimize the parameters and keep the model with the best perplexity on the validation set. During test time, we use a greedy search to generate text and calculate the BLEU-1,2,3 scores with the 5 references from the table. For the pre-trained models, we base our implementation on Huggingface's Transformer~\cite{Wolf2019HuggingFacesTS} for both BERT~\cite{devlin2019bert} and GPT-2~\cite{radford2019language} with subword unit vocabulary of 30K. During linearization, we found that using the whole table compromises the performance greatly, partly due to 1) over-length issue with pre-trained LM, 2) too much irrelevant information input. Therefore, we propose to use partial table as input, specifically, we run entity linking over the sentences to detect the linked columns of the table and only linearize the partial table as input $P_T$. 

Both are finetuned using Adam optimizer~\cite{kingma2014adam} with a learning rate of 1e-6. In both adversarial training and reinforcement learning algorithms, we add maximum likelihood objective to stabilize the training, we select the appropriate balancing factor based on the validation Adv-Acc socre. For coarse-to-fine training, we first warm up the model to generate the template sequence and then finetune it on the concatenated full sequence. Model selection is based on the bleu-3 score on validation split.

\begin{figure}[!t]
    \centering
    \includegraphics[width=1.0\linewidth]{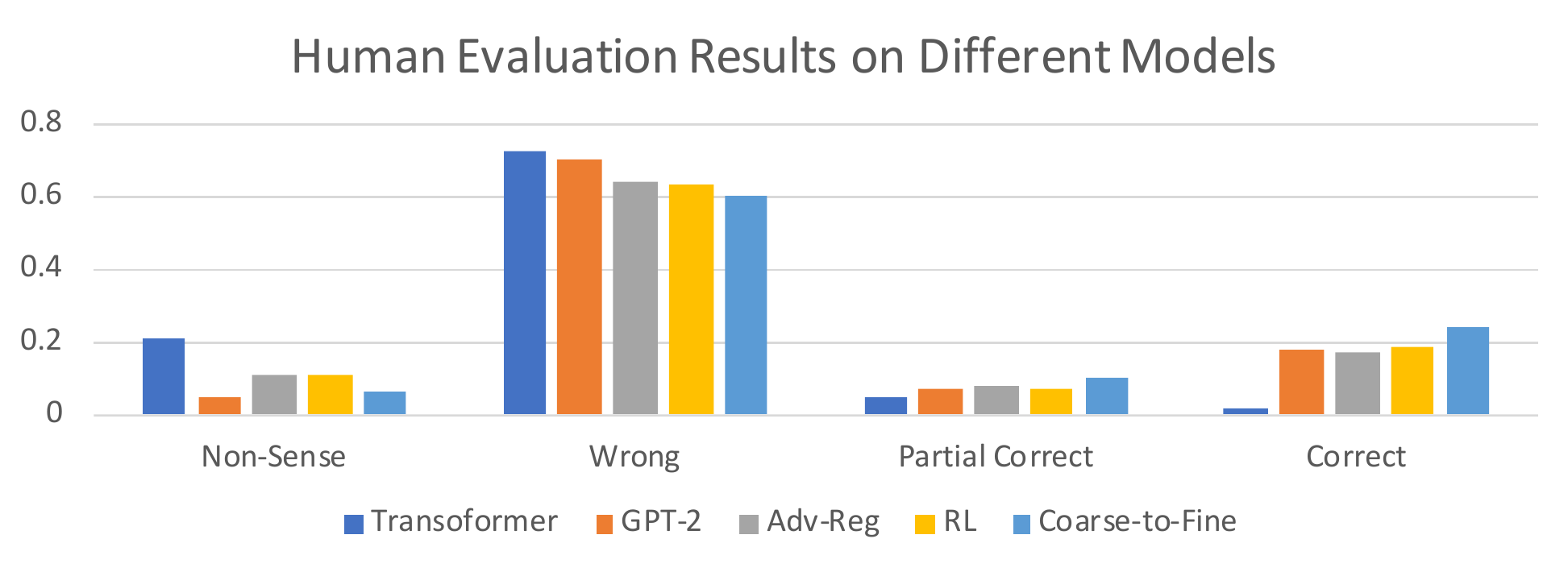}
    %\caption{The human evaluation results of different models on the sampled sentences.}
    \includegraphics[width=1.0\linewidth]{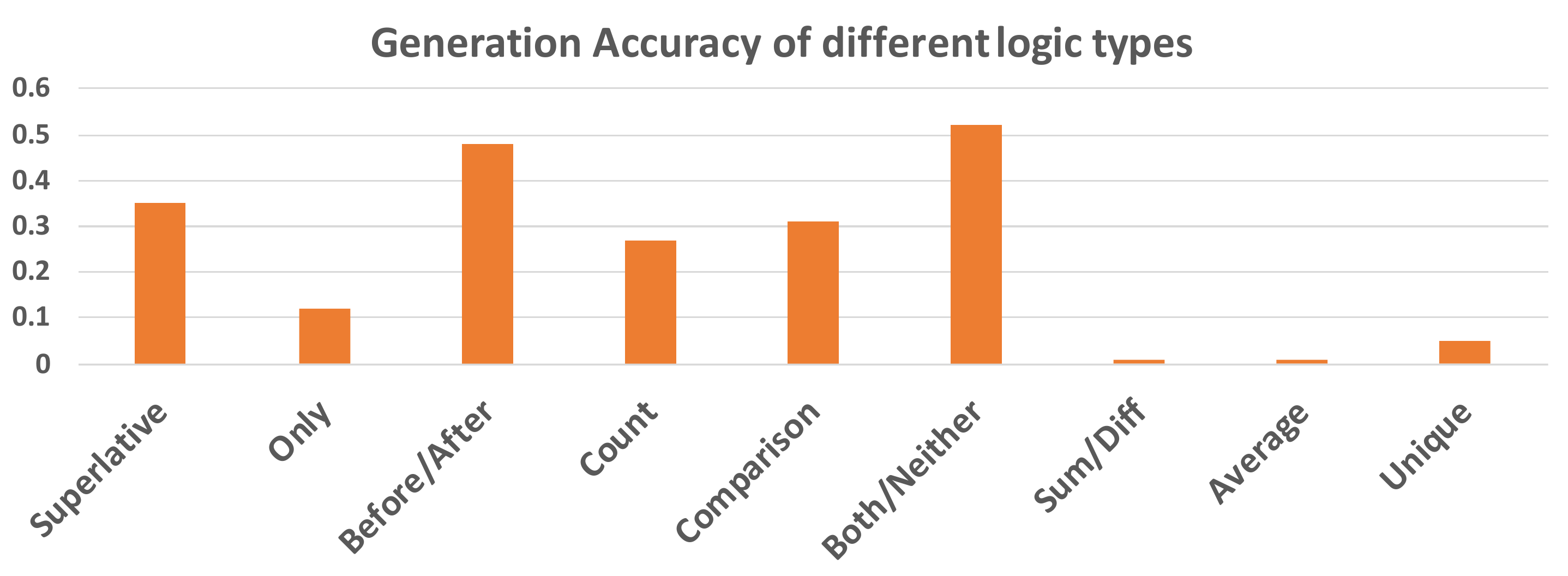}
    \caption{The human evaluation results of different models on the sampled sentences.}
    \label{fig:human}
    \vspace{-3ex}
\end{figure}
\subsection{Experimental Results}
We first perform an automatic evaluation to approximately measure the performance of different models and then conduct an in-depth human evaluation to have a better understanding. 
\paragraph{Automatic Evaluation:}
The experimental results are summarized in~\autoref{tab:un_pretrained}, where we comprehensively survey different architectures and training algorithms. For the non-pretrained models, we observe that Transformer is slightly better than LSTM and two different table encoding strategies achieve similar results. In contrast, pre-trained models are much better at lowering the perplexity, besides the generated sentences significantly outperform the non-pretrained models in terms of both fluency and fidelity score with GPT-TabGen and BERT-TabGen achieving similar performance. As the BERT-TabGen runs much slower due to multiple passes of decoding, we favor GPT-TabGen in the following experiments. With the adversarial regularization and reinforcement training, the model can only improve the optimized fidelity metric, with the fluency scores dropping significantly. Such phenomena confirm our assumption about the caveats of the monotonic generation paradigm. For the proposed coarse-to-fine generation scheme, as the ``[ENT]" tokens are replaced by entity names, which normally contain a phrase like ``Feb 2nd". Such n-gram phrase substitution preserves the completeness of entity names and thus leads to higher 2/3/4-gram matches, which translates to higher BLEU-3 and lower BLEU-1 in~\autoref{tab:un_pretrained}. The proposed coarse-to-fine generation can yield reasonable improvement over NLI-Acc and Adv-Acc, which demonstrates its advantages of in capturing logical dependency.
\paragraph{Human Evaluation}
To further investigate the quality of the generated text, we propose to perform human evaluation. Specifically, we sample 200 sentences from different models and distribute them independently to human experts (graduate students from the computer science department) to verify their quality. Specifically, the quality measure is categorized into categories: 1) non-sense: the sentence does not make much sense, which is mainly due to disfluency or repetition problem. 2) wrong: a fluent sentence with wrong logic. 3) partial-correct: the sentence contains more than one fact, at least one of them is correct 4) correct: the high-quality in both fluency and logic correctness. We demonstrate the results in~\autoref{fig:human}, from which we observe that pre-training significantly decreases the non-sense proportion. However, the RL and Adv-Reg both harm the fluency and lead to more non-sense sentences. In contrast, the coarse-to-fine model can maintain the non-sense proportion while significantly increasing correct/partial-correct sentences. From human evaluation, even the best performing model can get slightly over 20\% of its prediction logically correct, which reflects the challenges of \dataset for existing paradigm.
\paragraph{Evaluation Metrics}
We here analyze the effectiveness of the defined automatic evaluation metrics for fidelity evaluation. For the Parsing-based evaluation and NLI-based evaluation, we use the adversarial set (containing positive/negative sample pairs) to evaluate their consistency with human judges. Parsing-based model only achieves an accuracy of 60\%, while NLI-based model achieves a higher accuracy of 65\%. It indicates that the fidelity measurement model is itself a very challenging problem and the existing models are still in a premature stage. Therefore, the exact number of SP-Acc or NLI-Acc cannot reliably reflect the exact proportion of sentences logically entailed by the table. However, we still believe they are informative for model development based on the following reasons: 1) the automatic fidelity scores are quite stable, not sensitive to random initialization or different configurations, 2) when comparing different models (Transformer vs. GPT-2 vs. RL/Adv-Reg vs. Coarse-to-Fine), the trends of different automatic scores are consistent with human evaluation, which indicates its potential in assisting the development of new models.
\paragraph{Fine-grained Analysis}
To better understand the generation model's reasoning capability in regarding different logical operations, we pick the most frequent 9 operations (definition in the Appendix) and analyze the best model's capability in expressing these different logic. We demonstrate our human evaluation in~\autoref{fig:human} to make the following inspections: 1) the model performs best in justifying the order of different entities (before/after) and relating two entities (both/neither/comparison). 2) the model performs reasonably well at superlative and count operation.  3) the generation model performs much worse in operations like ``only, unique". 4) the model is not able to perform mathematical aggregation like average, sum, etc. Overall, the string-based operations are easier than numeric-based operations, how to infuse the numeric knowledge is an open research question to move forward.

%\begin{figure}[!t]
%    \centering
%    \includegraphics[width=1.0\linewidth]{human-evaluation-logic.pdf}
%    \caption{The human evaluation results of different models on the sampled sentences.}
%    \label{fig:human-logic}
%\end{figure}

%\subsection{Discussion}
%The non-pretrained models are more flexible since it can directly encode the table structure using a different network architecture, while the pre-trained models are more rigid since it has to linearize the table into the human language as input.

%\jianshu{I think the quantitative example analysis needs to do some sort of comparison between a few representative baselines. The key message of this paper is about logical factualness during generation: a dataset, new evaluation metrics, and existing methods' weakness in this aspect. Therefore, I think it would be more interesting to the reader if you can show different aspects of weakness for different existing baselines. Overall, the current paper is not strong enough in delivering a very clearcut takeaway message about this problem. Should reinforce this aspect.}

%\jianshu{Need human evaluation to justify the newly proposed evaluation metrics; it needs to show that the new metrics are aligned with human evaluation in terms of logical factualness or any other desired properties that are proposed earlier in this paper.}

\section{Related Work}
\paragraph{Natural Language Generation}
Natural language generation is a long-standing problem~\cite{kukich1983design,holmes1994text,reiter1997building}, which involves generating text from records or data. Recently, many neural-based generation models have been proposed~\cite{puduppully2019data,puduppully-etal-2019-data,lebret2016neural,wiseman2018learning} to achieve impressive performance on the existing datasets~\cite{chen2008learning,liang2009learning,lebret2016neural,dusek2019e2e,wiseman2017challenges} since the annotated text are mostly surface-level annotation without logical inference. Unlike them, \dataset has rich inference, which poses great challenges to existing models and evaluations.
\paragraph{Non-monotonic Generation}
There have been attempts recently to study the problem of non-monotonic text generation, which aims to teach the generation model to learn the generation order without external supervision~\cite{ford2018importance,welleck2019non,gu2019insertion,mansimov2019generalized}. These models have shown to learn rational generation order to approach similar performance as the left-to-right case. These approaches are useful at capturing more sophisticated dependency within the sentence, which provides a plausible direction to pursue in \dataset.
\paragraph{Factualness Evaluation}
Fidelity is an important research topic in generation, In ROTOWIRE~\cite{wiseman2017challenges} and MSCOCO~\cite{lin2014microsoft}, IE-based extractive evaluation~\cite{rohrbach2018object,dhingra2019handling} are adopted for surface-level matching to replace costly human evaluation. In abstractive summarization, \citet{goodrich2019assessing} proposes NER + Relation Classification method to investigate fidelity in generated summarization while \citet{kryscinski2019evaluating} proposes to use NLI models to understand the entailment between generated text with the given document. These evaluations are beyond surface-level to study more sophisticated linguistic phenomena like paraphrasing, compression, entailment, inclusion, etc, which are common in summarization tasks.

\section{Conclusion}
In this paper, we propose logical NLG to study the logical inference problem in generation. We conduct comprehensive experiments to show the existing NLG models are restricted by its monotonic nature and conclude this to be a proper next-step problem to study NLG systems. There are still some unsolved problems for Logical NLG, e.g. how to improve the quality of automatic metrics to better help human automatically judge models' performances. To promote the research in this direction, we host a LogicNLG challenge\footnote{\url{https://competitions.codalab.org/competitions/24527}} to help better benchmark the current progress.

\section{Acknowledgement}
The authors would like to thank the anonymous reviewers for their thoughtful comments.%, which greatly help them polish the paper.

\bibliography{acl2020}
\bibliographystyle{acl_natbib}

\appendix
\section{Dataset Examples}
\begin{figure*}[!h]
    \centering
    \includegraphics[width=0.85\linewidth]{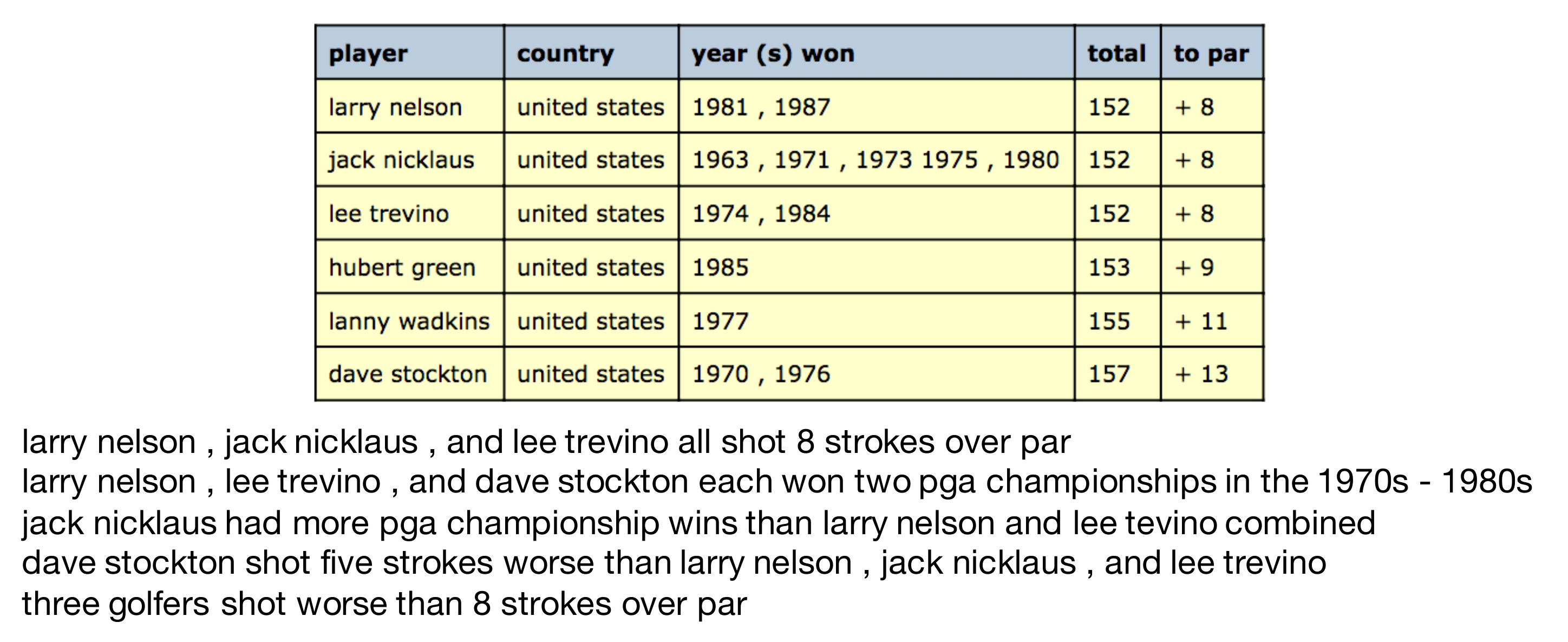}
    \caption{Example from \textsc{LogicNLG}.}
    \label{fig:ex1}
\end{figure*}

\begin{figure*}[!h]
    \centering
    \includegraphics[width=0.85\linewidth]{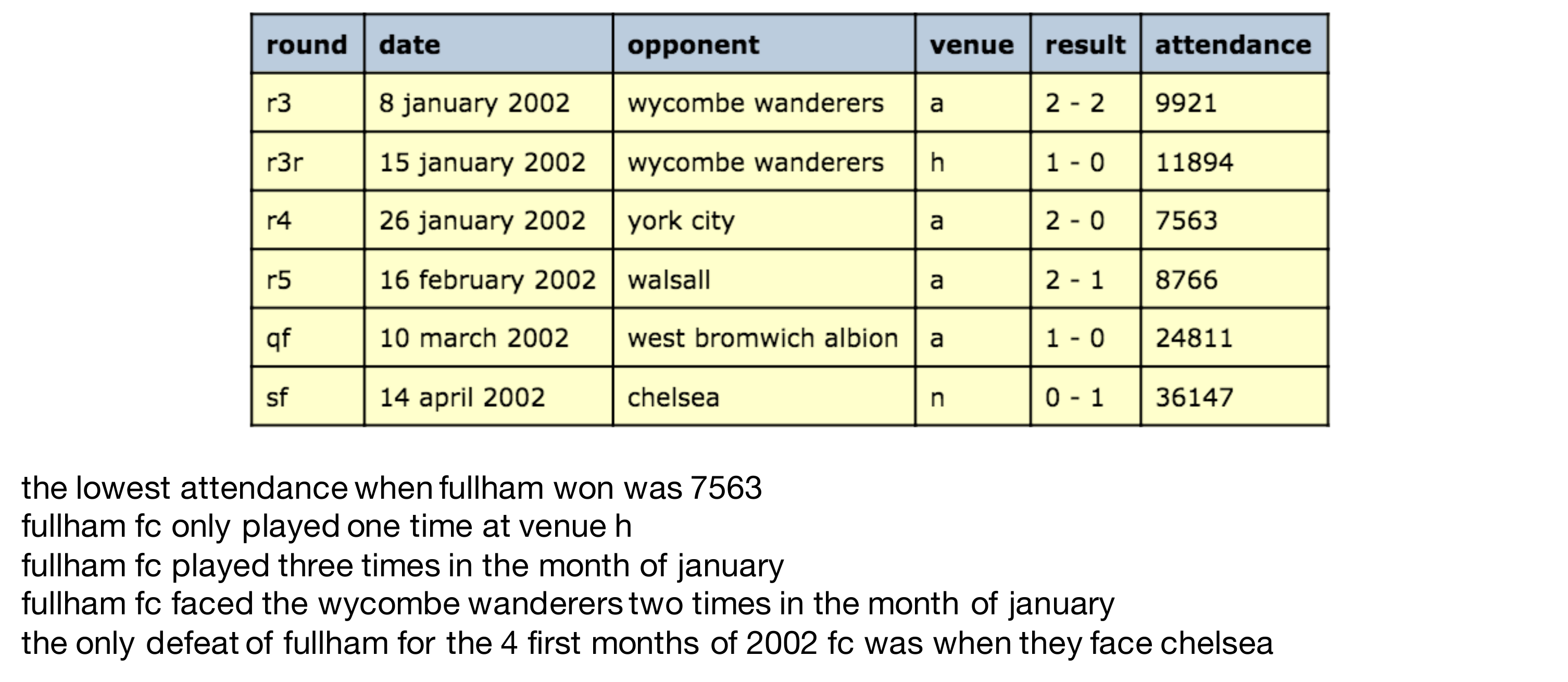}
    \caption{Example from \textsc{LogicNLG}.}
    \label{fig:ex2}
\end{figure*}

In order to give readers a better sense of the statements in \textsc{LogicNLG}, we demonstrate some typical examples below as~\autoref{fig:ex1} and~\autoref{fig:ex2}. Each table in the dataset is associated with five different examples covering diversified inference skills. For example, ~\autoref{fig:ex1} requires `all' operation to identify multiple rows having the same value on certain properties. ~\autoref{fig:ex2} requires the model to perform superlative, or count operation to identify the numerically highest number. 

\section{Logical Operation Distribution}
The dataset consists of the most common types of logical inference in our daily communication, to help the readers understand the semantic meaning of these inference, we list their definition and some examples below:
\begin{itemize}
    \item superlative: operations involving max,min or other comparison operation to get the lowest or highest value. Sentence: xxx is the tallest player in xxx team.
    \item only: operation to identify the single entity which has a unique property the other entries do not have. Sentence: xxx is the only person to win all the games.
    \item before/after: operations to compare time or spatial order. Sentence: xxx is born before xxx.
    \item count: operations to enumerate the amount of entries meeting certain criterion. Sentence: there are two people from the central united states.
    \item comparison: operations to compare two or given number of entities. Sentence: xxx has better income than xxx.
    \item both/neither: operations to summarize the common properties of two entries. Sentence: xxx and xxx are both from the same country.
    \item sum/diff: operations to perform numeric summation or difference between numbers. Sentence: xxx gives 1 more dollars than xxxx in the donation.
    \item average: the average number of people attending the game is 500.
    \item unique: the uniq operation in sql to assemble summarize different entities. Sentence: from the table, players are from 4 unique countries.
\end{itemize}
 
 \begin{figure*}[!t]
    \centering
    \includegraphics[width=1.0\linewidth]{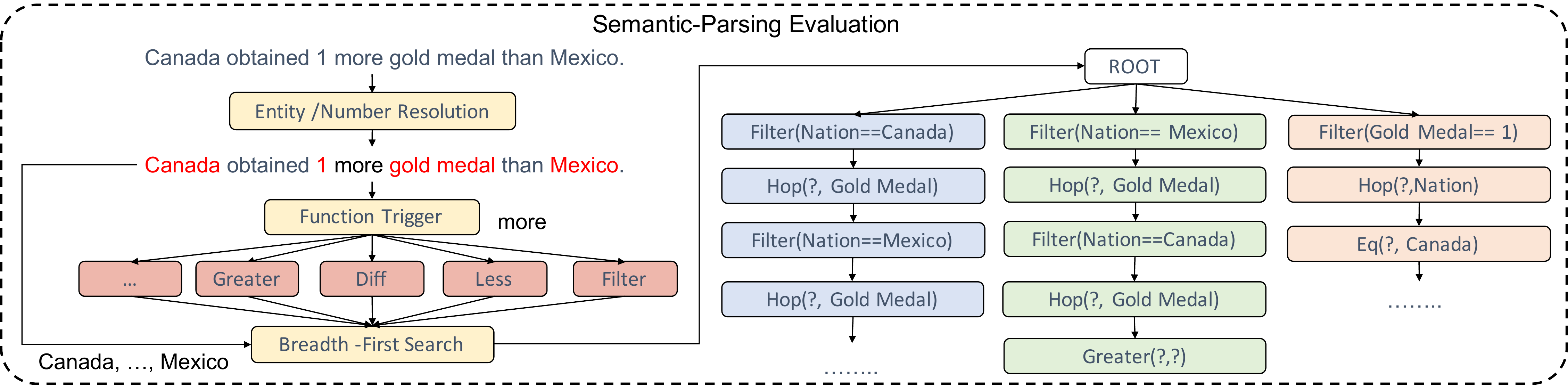}
    \caption{The BFS-based parser used in our evaluation.}
    \label{fig:parser}
\end{figure*}
\section{Semantic Parser}
Specifically, the scorer is realized by a matching model $f_{\gamma}$, which takes a logic form $P$ and the statement $Y$ to output a consistency score $f_{\gamma}(P, Y)$ between range of [0,1] with higher value indicating better consistency. As no groundtruth logical forms are provided, we utilize weakly supervised training. The set of logical forms generated is denoted as $\mathbf{P}$, the logical forms returning binary value of \texttt{True} is viewed as pseudo positive example $\mathbf{P}^+$ and the logical forms returning \texttt{False} is treated as pseudo negative example $\mathbf{P}^-$. We propose to optimize the following objective to discriminate two sets:
\begin{gather*}
\small
\begin{split}
\argmax_{\gamma} \expect{(\textbf{T}, Y) \in D_{train}} [\expect{P \in \mathbf{P}^+} [f_{\gamma}(P, Y)] -  \expect{P \in \mathbf{P}^-} [f_{\gamma}(P, Y)]]
\end{split}
\end{gather*}
As demonstrated in~\autoref{fig:parser}, our semantic parser is comprised of three different parts, namely a resolution model, a breadth-first search model and a ranker model. The resolution model will try to figure out what are the entities appearing in the table and what are the numbers it needs to infer. These results are pushed to a buffer as the initial point, then the BFS search will try to compose plausible logical forms based on the values from the buffer. However, most of the synthesized logical forms are not relevant to the semantics the sentence is aimed to express. In the end, we need to train a ranker, which can learn to identify the most consistent logical form and use that to represent the formal semantics of given sentence.  

\section{Qualitative Example}
\begin{figure*}[!t]
    \centering
    \includegraphics[width=0.85\linewidth]{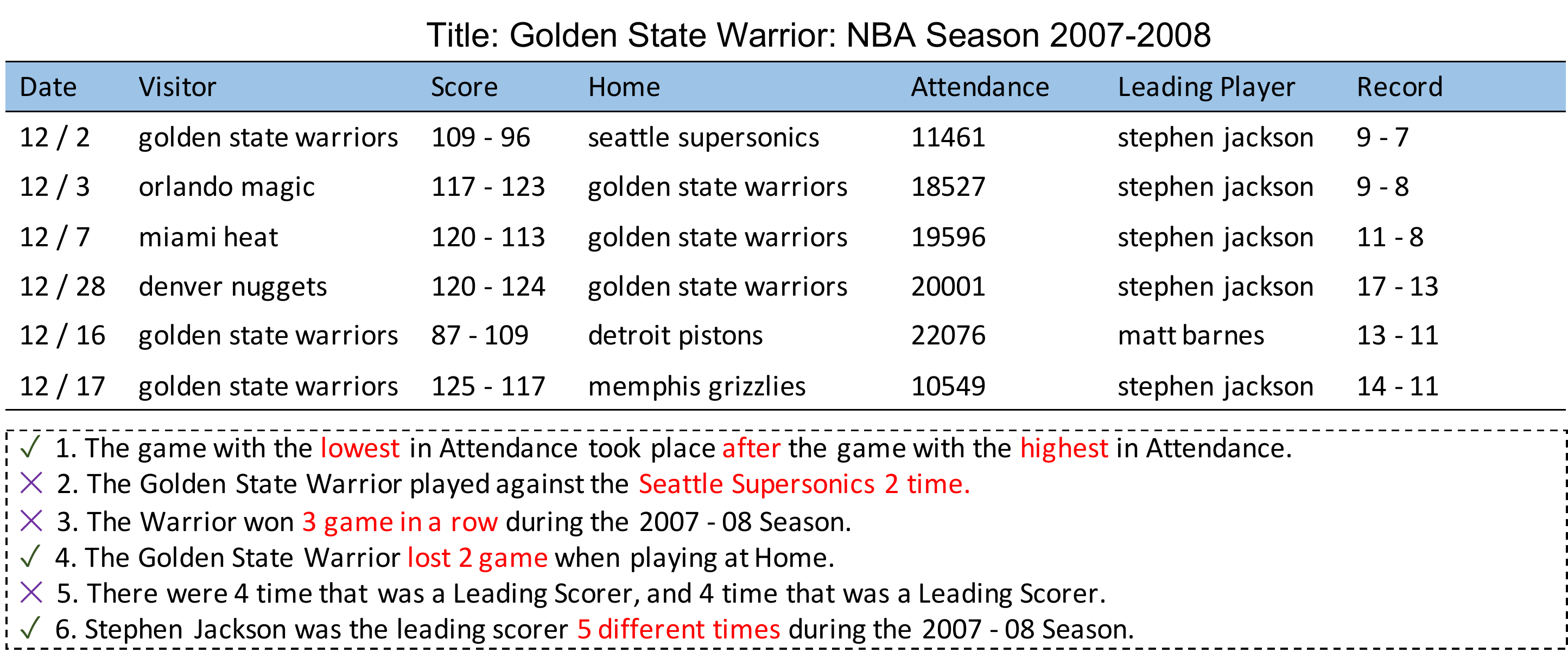}
    \caption{The statements generated by GPT-TabGen model with random sampling.}
    \label{fig:quantitative}
\end{figure*}
Next, we demonstrate some generated samples in~\autoref{fig:quantitative}, which are generated from a table crawled from Wikipedia page\footnote{\url{https://en.wikipedia.org/wiki/2007\%E2\%80\%9308_Golden_State_Warriors_season}}. Though most of the text generated by the model is coherent and reasonable, we do observe some disfluency like repetition, contradiction, erroneous sentences like the sentence 5. For the other sentences, three of them are logically correct, the first sentence contains quite complex logic with three different symbolic operations ``argmax, argmin, after". The fourth and sixth sentences involve operations like ``filter, count". In contrast, the second and third examples are factually incorrect as the team only competes with ``Seattle" once and the 3 games are not in a row. We can see that the errors are quite diversified, it is difficult to debug what is the source of these errors by simply looking into the deep generation model. In the future, more interpretable generation model need to be built to make the inference process more transparent. 

\end{document}